\documentclass{llncs}

\usepackage[a4paper,includeheadfoot,top=1.65in,bottom=1.65in,left=1.73in,hcentering]{geometry}
\usepackage[pdftex]{graphicx}
\usepackage[sectionbib]{natbib}
\usepackage{caption}
\usepackage{adjustbox}
\usepackage{makecell} 
\usepackage{booktabs}
\usepackage{tikz}

\usepackage[utf8]{inputenc}
\usepackage{t1enc}
\usepackage{siunitx}
\usepackage{tikz-dependency}
\usepackage{float}
\restylefloat{figure}
\usepackage{tikz-qtree}
\usepackage{hyperref}
\usepackage{multirow}
\usepackage{siunitx}     
\usepackage{caption}     
\usepackage{array}       
\usepackage{ulem}  
\usepackage{tabularx}    
\usepackage{makecell}    
\usepackage{geometry}    

\renewcommand\bibname{\refname}

\newif{\ifhidecomments}
\hidecommentsfalse

\ifhidecomments
    \newcommand{\botond}[1]{}
    \newcommand{\ham}[1]{}
    \newcommand{\judit}[1]{}
\else
    \newcommand{\botond}[1]{\textcolor{red}{[#1 ({\bf Botond})]}}
    \newcommand{\ham}[1]{\textcolor{blue}{[#1 ({\bf Endre})]}}
    \newcommand{\judit}[1]{\textcolor{orange}{[#1 ({\bf Judit})]}} 
\fi

\begin{document}

\title{HuAMR: A Hungarian AMR Parser and Dataset}
\author{Botond Barta\inst{1}, Endre Hamerlik\inst{1}, Mil\'an Konor Nyist\inst{2}, Judit \'Acs\inst{1}\\
\institute{
$^1$HUN-REN Institute for Computer Science and Control\\
$^2$Eötvös Loránd University}
\email{\{botondbarta,hamerlik.endre,acsjudit\}@sztaki.hu}, \email{nyist.milan78@gmail.com}
}

\maketitle

\begin{abstract}
We present \textbf{HuAMR}, the first Abstract Meaning Representation (AMR) dataset and a suite of large language model-based AMR parsers for Hungarian, targeting the scarcity of semantic resources for non-English languages. To create HuAMR, we employed Llama-3.1-70B to automatically generate silver-standard AMR annotations, which we then refined manually to ensure quality. Building on this dataset, we investigate how different model architectures — mT5 Large and Llama-3.2-1B — and fine-tuning strategies affect AMR parsing performance.

While incorporating silver-standard AMRs from Llama-3.1-70B into the training data of smaller models does not consistently boost overall scores, our results show that these techniques effectively enhance parsing accuracy on Hungarian news data (the domain of HuAMR). We evaluate our parsers using Smatch scores and confirm the potential of HuAMR and our parsers for advancing semantic parsing research.
\end{abstract}

\section{Introduction}

Large Language Models (LLMs) have revolutionized the field of natural language processing, achieving state-of-the-art results across tasks like translation, summarization, and question answering \citep{brown2020language,openai2023gpt4,touvron2023llama}. These models have demonstrated remarkable capabilities in understanding and generating human-like text, opening up a wide range of research and application possibilities.

Despite these advancements, LLMs often exhibit issues related to factual accuracy and consistency, sometimes generating outputs that are incorrect or misleading \citep{ji2023survey,maynez2020faithfulness}. This limitation is particularly evident in tasks requiring deep semantic understanding and faithful representation of information \citep{cao2022hallucinated}.

A critical challenge underlying these issues is the representation of meaning in the generated text. LLMs primarily rely on statistical patterns in data, which can lead to a lack of deep semantic comprehension \citep{bender2021dangers}. Abstract Meaning Representation (AMR) \citep{banarescu2013abstract} offers a solution through a structured, graph-based formalism that represents the core meaning of sentences. This representation enables more accurate language understanding and generation.

AMR has been successfully applied in various contexts to improve semantic interpretation. For instance, \citet{zhu2020incorporating} explored ways to integrate AMR-derived graph information into Transformer-based architectures, demonstrating how semantic graphs can enhance model performance on parsing tasks. \citet{dou2022gsum} further showed that leveraging AMR can improve factual consistency in summarization, underscoring the potential of AMR to enhance both the quality and accuracy of generated text.

Despite these advancements, most AMR resources and tools are available only for English, limiting the broader applicability of these techniques \citep{blloshmi2020xl}. There is a critical need to develop AMR tools for a wider range of languages to enhance the semantic capabilities of LLMs globally.

In this work, we address this gap by translating the gold standard AMR (AMR 3.0 \citet{amr3}) dataset into Hungarian, creating a valuable resource for this language. We present and publish a suite of language models, ranging from small (1B parameters)  to large (70B parameters), capable of generating AMRs for Hungarian texts\footnote{Resources available at \url{https://github.com/botondbarta/HuAMR}}. Our main contributions are:

\begin{itemize}
    \item \textbf{Creation of a Hungarian AMR dataset:} We translate the AMR 3.0 dataset into Hungarian, providing the first AMR resource for this language. We also publish a synthetic AMR dataset (HuAMR) generated with a Llama-3.1-70B model for further coverage.
    \item \textbf{Development of high-performance Hungarian AMR parsers:} We develop and release a series of AMR parsers for Hungarian texts, via finetuning mT5 Large and Meta Llama 3.2 models. These models enable accurate AMR graph generation for Hungarian, a significant advancement for low-resource language processing.
    \item \textbf{Extensive evaluation of modeling techniques:} We conduct a thorough analysis of different model architectures, examining the impact of additional silver training data on AMR parsing performance. Our evaluation provides valuable insights into optimizing cross-lingual AMR parsers for non-English languages.
    \item \textbf{Evaluation of data augmentation strategies:} We leverage larger models to generate silver-standard AMR annotations for Hungarian texts, which are then incorporated into the training data for smaller models. This approach enhances the performance of smaller models, demonstrating an effective strategy for low-resource settings.
\end{itemize}

These efforts pave the way for extending AMR parsing to multiple languages and contribute toward the development of multilingual AMR parsers.

\section{Abstract meaning representation}
\begin{figure}
	\begin{minipage}[t]{0.45\textwidth}
		\centering
		  Graph representation\\
		\vspace{0.5cm}
		\includegraphics[scale=0.7]{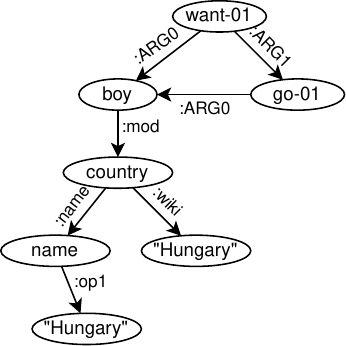}
	\end{minipage}
	\begin{minipage}[t]{0.5\textwidth}
		\centering 
		PENMAN notation \\
		\vspace{0.1cm}
		\begin{verbatim}
    (w / want-01
       :ARG0 (b / boy
            :mod (c / country
                :wiki "Hungary"
                :name (n / name
                    :op1 "Hungary")))
       :ARG1 (g / go-01
            :ARG0 b))
            \end{verbatim}
	\end{minipage}

        \begin{minipage}[t]{1\linewidth}
            \vspace{0.5cm}
            \centering
            Depth-first serialization after the preprocessing steps described in Section \ref{sec:preprocess}.
            \begin{verbatim}
( w / want-01 :ARG0 ( b / boy :mod ( c / country :name ( n / name :op1
"Hungary" ) ) ) :ARG1 ( g / go-01 :ARG0 b ) )
            \end{verbatim}
        \end{minipage}
	\caption{AMR notations for the sentence "The Hungarian boy wants to go".}
	\label{fig:preprocess}
\end{figure}

AMR is a semantic framework that represents the meaning of a sentence using a graph structure, abstracted from specific lexical and syntactic forms. It captures key relationships, such as who performs an action, what the action is, and who or what it affects. In an AMR graph, nodes represent concepts like verbs, entities, and PropBank framesets \citep{kingsbury2002treebank}, while edges denote semantic roles and relationships between these concepts. In cross-lingual setting the nodes and edges are always labeled in English regardless of the input language, and the entities are always lemmatized. AMR graphs are composed of logical triples that describe relationships between sources and targets. These triples provide an abstract, computationally readable representation of meaning, which can be expressed in both graph and PENMAN notation, allowing AMR to be applied in various natural language processing tasks like machine translation and information extraction. An example for the sentence \textit{"The Hungarian boy wants to go"} is shown in Figure~\ref{fig:preprocess} using graph representation and PENMAN notation. The sentence is represented with variables and concepts like \texttt{want-01} and \texttt{go-01}, linked to the subject (\textit{"the boy"}) and the action. Modifiers like the nationality "Hungarian" are captured using the \texttt{:mod} role, while the \texttt{:wiki} role links the modifier to its corresponding Wikipedia page, "Hungary". The official and more detailed description of AMR is available in the AMR specification\footnote{\url{https://github.com/amrisi/amr-guidelines/blob/master/amr.md}}.

\section{Related Work}

\citet{vanroy2024less} explored cross-lingual AMR parsing for English, Dutch, and Spanish using both multilingual and monolingual configurations of BART-large. Their findings indicate that monolingual models consistently outperform multilingual ones, highlighting the limitations of multilingual approaches in this context.

Instruction fine-tuned language models show strong generalization to unseen tasks \citep{chung2024scaling,ouyang2022training}. However, standard parsing tasks like AMR are usually excluded from such instruction datasets and models. \citet{lee2023amr} experimented with FLAN-T5 instruction fine-tuned models showing significant improvement in AMR parsing over earlier BART \citep{lewis-etal-2020-bart} based models. Their work included a two step training approach: first a full fine-tuning and then a parameter efficient fine-tuning using Low Rank Adaptation (LoRA) \citep{hu2021lora}.

\citet{regan-etal-2024-massive} created MASSIVE-AMR, a cross-lingual AMR dataset in the domain of question answering (QA) consisting of 1,685 utterances which are localized in 52 languages, Hungarian included. However the dataset contains some problems, as entities are inconsistently translated or left untranslated, and sometimes appear in inflected forms, making the parsing process difficult. It also lacks punctuation marks at the end of sentences, which are crucial as they often indicate parts of the AMR.

More recently, \cite{kang2024should} experimented with cross-lingual AMR parsing across 13 languages using a meta-learning model. They found that fine-tuning a model on a small number of examples from previously unseen languages failed to improve parser performance, underscoring the challenges of cross-lingual generalization in AMR parsing.

\section{Data}
\subsection{Gold AMR graph - silver translation}
To create Hungarian training data from the AMR 3.0 dataset we translated each sentence into Hungarian using DeepL\footnote{\url{https://www.deepl.com/}}, keeping the AMRs unchanged. We refer to this translated dataset as AMR$^{\text{trans}}$. We evaluated the translation quality with a reference-free evaluation metric, COMET \citep{rei-etal-2020-comet}. The COMET score of the translation is $85.2\pm0.7$.

\subsection{Silver Training Data}
Following the methodology of \cite{damonte-cohen-2018-cross} we used the Europarl parallel corpus \citep{koehn-2005-europarl} to generate silver training data. We used an English AMR parser\footnote{\url{https://github.com/bjascob/amrlib-models/releases/parse_xfm_bart_large-v0_1_0}} to produce silver quality AMR graphs for the English sentences. These AMR graphs then served as target representations for their corresponding Hungarian translations.

We also experimented with creating another silver training dataset by first fine-tuning an adapter on the Llama-3.1-70B\footnote{\url{https://huggingface.co/meta-llama/Llama-3.1-70B-Instruct}} model using the AMR$^{\text{trans}}$ dataset. The exact training process for this model is described in Section \ref{sec:silver-model}. With this adapted model, we generated AMR graphs for 50k samples drawn from the Hunsum-2 corpus \citep{barta-etal-2024-news} using the first sentence of each selected article’s lead. We named this dataset \textit{HuAMR}.

To ensure that the generated silver AMR graphs are high-quality, we validated them against PropBank frame argument descriptions verifying that the argument structure of the frames in the graphs matched the arguments specified in PropBank. Many of the generated graphs included unnecessary '\texttt{and}' operators. These operators were often introduced as the first node in the graph, but the model frequently failed to end the AMR properly, resulting in incomplete structures. This problem broke the logical flow of the AMR, making the graph inconsistent with the meaning of the sentence. Therefore, we validated the graphs to ensure that whenever an '\texttt{and}' is used, it has at least two operands. With these validations we discarded 6189 instances from the HuAMR dataset. Of the remaining data, 3811 has been reserved for testing and the remaining 40k are available for training.

We compared the three different datasets by the 15 most frequent top nodes. These statistics are shown in Table \ref{tab:frequent-nodes}. Notably, the '\texttt{and}' concept appears as the most frequent node across all three datasets. Other frequently appearing nodes include '\texttt{say-01}' and '\texttt{contrast-01}' which are prevalent in AMR 3.0 and HuAMR, but different nodes like '\texttt{obligate-01}' and '\texttt{recommend-01}' appear more frequently in Europarl.

It is important to note that HuAMR consists exclusively of news data, whereas only a small fraction of AMR 3.0 is derived from news articles. This difference in data composition may explain the variations in node frequency, as news data typically includes a distinct set of linguistic structures and terminology compared to other types of content.

\begin{table}[!htbp]
\centering
\begin{tabular}{lc|lc|lc}
\toprule
\textbf{AMR 3.0} & \textbf{\#} & \textbf{HuAMR}  & \textbf{\#} & \textbf{Europarl} & \textbf{\#} \\ \midrule
and             & 7.0k    & and             & 4.7k      & and               & 57.7k        \\
say-01          & 3.0k    & say-01          & 4.6k      & contrast-01       & 37.4k        \\
contrast-01     & 3.0k    & contrast-01     & 2.2k      & say-01            & 34.5k        \\
multi-sentence  & 1.7k    & possible-01     & 1.9k      & obligate-01       & 24.0k        \\
possible-01     & 1.7k    & state-01        & 0.7k      & cause-01          & 23.5k        \\
cause-01        & 1.6k    & cause-01        & 0.7k      & possible-01       & 19.4k        \\
state-01        & 1.5k    & have-concession & 0.6k      & recommend-01      & 15.8k        \\
have-concession & 0.9k    & win-01          & 0.5k      & multi-sentence    & 14.0k        \\
think-01        & 0.9k    & announce-01     & 0.4k      & need-01           & 11.2k        \\
person          & 0.7k    & find-01         & 0.4k      & like-02           & 10.0k        \\
have-03         & 0.6k    & report-01       & 0.3k      & have-concession   & 8.8k         \\
have-condition  & 0.6k    & show-01         & 0.3k      & believe-01        & 7.4k         \\
date-entity     & 0.5k    & write-01        & 0.3k      & think-01          & 6.7k         \\
know-01         & 0.5k    & start-01        & 0.3k      & important-01      & 5.4k         \\
have-degree     & 0.4k    & decide-01       & 0.2k      & have-03           & 4.7k         \\ 
\bottomrule
\end{tabular}
\caption{The top-15 most frequent AMR nodes in AMR 3.0, HuAMR and Europarl.}
\label{tab:frequent-nodes}
\end{table}

\subsection{Preprocessing}
\label{sec:preprocess}
Before training we remove the wiki tags from the AMR graphs. The graphs are then serialized using a depth-first approach, replacing newlines with spaces and compressing multiple spaces into a single one. To ensure consistent tokenization, a space is inserted before and after each parenthesis, as some tokenizers may otherwise interpret consecutive parentheses as a single token. An example serialization process is shown in Figure \ref{fig:preprocess}.

\section{Experiments}
\subsection{Model for silver data generation} \label{sec:silver-model}
We fine-tuned a Llama-3.1-70B-Instruct model using LoRA under a 4-bit quantization setup to enable efficient training on limited computational resources as it reduces the memory and compute requirements while preserving much of the model’s capabilities, which makes it possible to train such a large model. LoRA allows us to fine-tune the model by introducing a small number of trainable parameters in low-rank matrices, significantly reducing computational overhead compared to full fine-tuning.  In our setup, we applied LoRA to all projection layers of the transformer architecture. The LoRA specific hyperparameters included a rank of 8, an alpha value of 8 with a learning rate of 5e-5. The task type was set to \texttt{CASUAL\_LM}.

\subsection{Fine-tuning with additional silver data}
We explored the effect of varying the amount of silver data on model performance by training multiple models. Our experiments were carried out using two distinct architectures: the mT5-large model \citep{xue-etal-2021-mt5} that follows a sequence-to-sequence approach and the Llama-3.2-1B-Instruct model, reflecting the recent trend toward decoder-only architectures. Both models are within the 1.2-1.3 billion parameter range.

For training, we used subsets from the training set of both the HuAMR corpus and the silver Europarl corpus, testing different amounts of silver data in each experiment. From the HuAMR corpus, we experimented with four data sizes: 10k, 20k, 30k, and 40k samples. These subsets were combined with the AMR$^{\text{trans}}$ dataset to assess how increasing portions of silver data affected performance.

We also trained models using data from the silver Europarl corpus with 40k, 160k and 520k additional sentences. This allowed us to investigate how the models respond to different amounts of data from a parallel corpus created by a high-quality parser.

\section{Results}
\subsection{Silver data}
We evaluated the performance of our models on the test set of AMR$^{\text{trans}}$ and HuAMR using Smatch scores \citep{cai-knight-2013-smatch}. The Smatch score measures the semantic similarity between two AMR graphs by calculating the overlap of matched triples. For these calculations we used the Smatchpp \citep{opitz-2023-smatch} library. The scores are shown in Table \ref{tab:model_smatch}.

\begin{table}[!ht]
\centering

\begin{tabular}{lcc|cc}
\toprule
\multirow{2}{*}{\textbf{Model}} & \multicolumn{2}{c|}{\textbf{~~~~~~~~Additional Data~~~~~~~~}} & \multicolumn{2}{c}{\textbf{~~~~~~~~Smatch F$_1$~~~~~~~~}} \\
\cmidrule(lr){2-3} \cmidrule(lr){4-5}
                                & {~~~~HuAMR~~~~} & {~~~~Europarl~~~~} & {~~~~AMR$^{\text{trans}}$~~~~} & {~~~~HuAMR~~~~} \\ 
\midrule
\multicolumn{5}{c}{mT5 Large} \\
\midrule
mT5 Large & -     & -     & \textbf{\underline{72.90}} & 69.32 \\
mT5 Large & 10k   & -     & 72.28           & 73.65 \\
mT5 Large & 20k   & -     & 72.59           & 75.99 \\
mT5 Large & 30k   & -     & 71.98           & 75.26 \\
mT5 Large & 40k   & -     & 72.22           & \textbf{\underline{76.11}} \\
\midrule
\multicolumn{5}{c}{Llama 3.2 1B} \\
\midrule
Llama 3.2 1B & -     & -     & 67.43           & 54.44 \\
Llama 3.2 1B & 10k   & -     & 68.35           & 67.65 \\
Llama 3.2 1B & 20k   & -     & 68.41           & 70.10 \\
Llama 3.2 1B & 30k   & -     & 69.44           & 71.98 \\
Llama 3.2 1B & 40k   & -     & \textbf{69.59}           & \textbf{73.03} \\
\midrule
\midrule
Llama 3.2 1B & -     & 40k   & 68.67           & 60.05 \\
Llama 3.2 1B & -     & 160k  & \textbf{70.70 }          & \textbf{62.92} \\
Llama 3.2 1B & -     & 520k  & 66.16           & 59.08 \\
\bottomrule
\end{tabular}

\caption{Smatch F$_1$ scores for Llama 3.2 1B and mT5 Large models with increasing training data sizes from silver datasets HuAMR and Europarl. A dash (`-`) indicates no additional data. The best score for each model is \textbf{bolded}, and the overall best score is both \textbf{\underline{bolded and underlined}}.}
\label{tab:model_smatch}
\end{table}

Table \ref{tab:model_smatch} reveals several key trends:
\begin{itemize}

\item \textbf{Performance Superiority of mT5 Large:} Across all configurations of additional silver data from HuAMR and Europarl, mT5 Large consistently outperforms Llama 3.2 1B in Smatch scores.
\item \textbf{Saturation on AMR$^{\text{trans}}$ Evaluation:} Both models exhibit saturated performance on the AMR$^{\text{trans}}$ dataset; additional silver data yields minimal improvements in Smatch scores. This suggests that the models may have already reached their maximum potential for this particular dataset, and additional data may not provide sufficient new information to enhance their parsing accuracy.

\item \textbf{Llama 3.2 1B's Limitations with Europarl Data:} Incorporating up to 520k additional Europarl training instances benefits Llama 3.2 1B, but this model does not achieve parity with mT5 Large’s performance.
\end{itemize}

\begin{figure}
\centering
\includegraphics[width=0.9\textwidth]{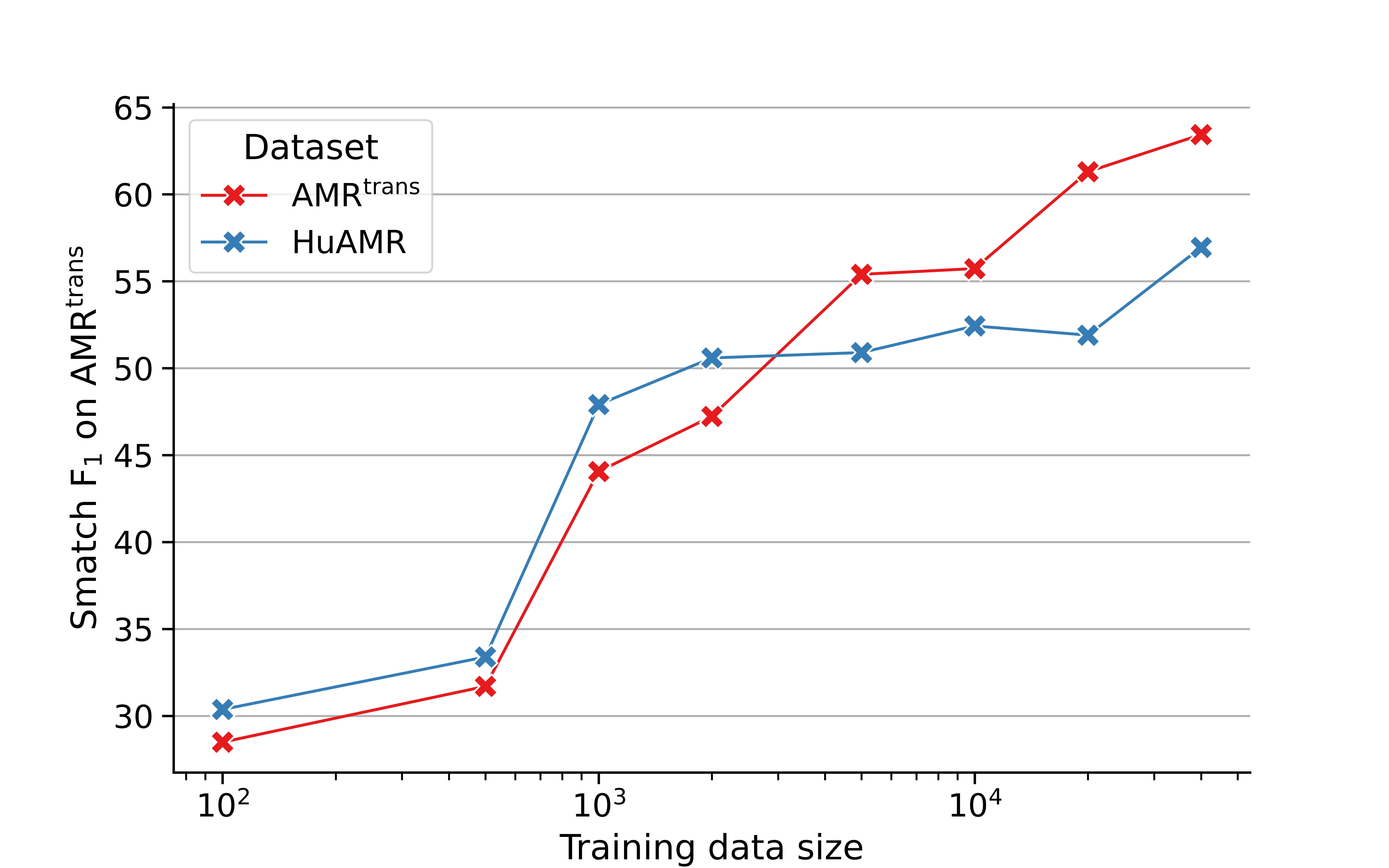}
\caption{Smatch F$_1$ score as a function of the training data size.}
\label{fig:training_curve}
\end{figure}

Building on the observation of performance saturation on the AMR$^{\text{trans}}$ test set despite increasing amounts of silver data, we conducted a control experiment to investigate whether the mT5 Large model is already saturated on the training set of AMR$^{\text{trans}}$. We trained the models using varying amounts of both AMR$^{\text{trans}}$ and HuAMR training data while evaluating performance on the AMR$^{\text{trans}}$ test set. The results, shown in Figure~\ref{fig:training_curve}, reveal that introducing a small amount of HuAMR data (100–2000 data points) provides a slight positive shift in performance. However, when scaling the AMR$^{\text{trans}}$ data from 4000 to 40k data points, models trained on AMR$^{\text{trans}}$ data obtain a consistent performance advantage of roughly 5\% in Smatch F$_1$ scores over those augmented with HuAMR. These trends confirm that the models are not overfitted or saturated when trained solely on 40k AMR$^{\text{trans}}$ data points.

\section{Conclusion}

This study presents the first Abstract Meaning Representation (AMR) resource for Hungarian by translating the gold-standard AMR 3.0 dataset \citep{amr3}. Utilizing this resource, a suite of AMR parsers for Hungarian texts was developed, leveraging models ranging from 1B to 70B parameters. The models demonstrate accurate AMR graph generation capabilities for Hungarian, significantly advancing semantic parsing for this underrepresented language.

We conducted an extensive evaluation of modeling techniques to examine the impact of additional silver training data and model size on AMR parsing performance. The results indicate that the mT5 Large model consistently outperforms the Llama 3.2 1B model across all configurations of additional silver data from HuAMR and Europarl. Despite substantial additions of silver data, both models showed limited improvements on the AMR$^{\text{trans}}$ test set.

Our control experiment indicates that increasing the quantity of gold-standard data leads to continued improvements in Smatch F$_1$ scores. Models trained on 40k AMR$^{\text{trans}}$ data points achieved approximately 5\% higher scores than those trained on the silver standard HuAMR.

These findings suggest that the models did not overfit nor saturate when trained solely on the full AMR$^{\text{trans}}$ dataset. The limited impact of additional silver data implies that the performance saturation is attributable to the lower quality of silver-standard annotations. The results highlight the critical role of high-quality training data and adequate model capacity in achieving optimal performance in semantic parsing tasks.

\section*{Acknowledgements}

This study was supported by the European Union project RRF-2.3.1-21-2022-00004 within the framework of the Artificial Intelligence National Laboratory, Hungary.

%
\renewcommand\bibname{References}
\bibliographystyle{splncsnat_en}
\bibliography{mszny}

\end{document}